%% file: main.tex
\documentclass[runningheads]{llncs}

\usepackage{eccv}

\usepackage{eccvabbrv}

\usepackage{graphicx}
\usepackage{booktabs}
\usepackage{bigfoot}

\usepackage[accsupp]{axessibility}  

\usepackage{hyperref}
\hypersetup{hidelinks}

\usepackage{orcidlink}

\makeatother

\begin{document}

\title{RegHead: Non-Humanoid Head Blendshapes via Feed-Forward Registration} 

\titlerunning{RegHead}

\author{Jiahao Luo\inst{1}\and
Hao Zhang\inst{2} \and
Jianqi Chen\inst{3} \and
Yijie He\inst{4} \and
Jiaxu Zou\inst{4} \and
\\
Michael Vasilkovsky\inst{4} \and
Sergei Korolev\inst{4} \and
Sergey Tulyakov\inst{4} \and
Chaoyang Wang\inst{4} \and
Peter Wonka\inst{3,4} \and
James Davis\inst{1} \and
Jian Wang\inst{4}
}

\authorrunning{J. Luo et al.}


\institute{
$^{1}$UCSC \quad
$^{2}$UIUC \quad
$^{3}$KAUST \quad
$^{4}$Snap~Inc.
}

\maketitle

\begin{abstract}
We present RegHead, a framework for constructing semantic blendshape sets for animatable non-humanoid head avatars. With a fixed expression vocabulary, semantic blendshapes provide a low-dimensional and interpretable animation interface and support cross-identity retargeting. Building such blendshape sets remains expensive because (i) expression-consistent supervision is scarce, (ii) generated 4D assets typically lack correspondence, and (iii) facial motion is highly localized. We propose (1) a large-scale dataset of non-humanoid identities paired with a shared expression vocabulary, obtained by expanding a small artist-rigged library via fine-tuned image editing; (2) a dense stochastic anchor motion representation tailored to localized facial deformations; and (3) a fast feed-forward registration model that converts unregistered expression meshes into a corresponded blendshape basis by predicting anchor-based deformations from the neutral shape. Experiments show that our approach produces higher-fidelity expression meshes than baselines, while running orders of magnitude faster than optimization. We further demonstrate real-time retargeting from human face tracking signals to non-humanoid characters, capturing both head pose and localized facial motions. Our project page is available at \url{https://snap-research.github.io/RegHead/}.
  \vspace{-7mm}
\end{abstract}

\begin{figure*}[t]
\centering
\includegraphics[width=\textwidth]{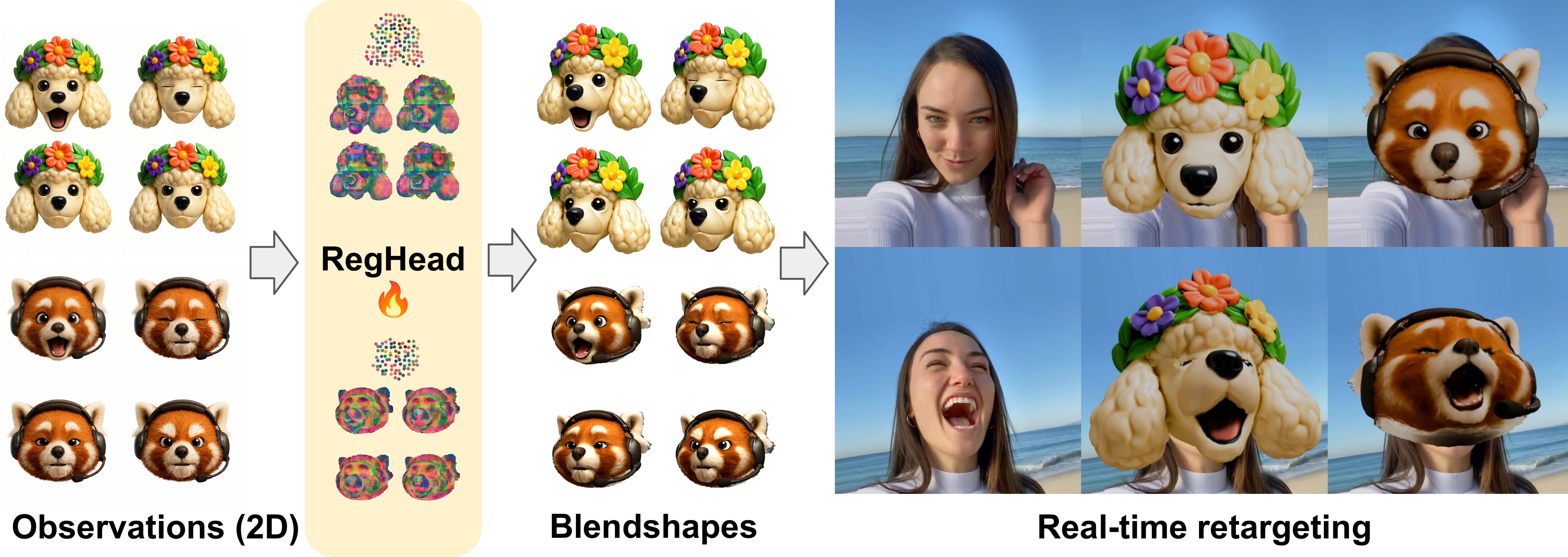}
\caption{
\textbf{RegHead} converts semantically labeled expression observations into a corresponded semantic blendshape set for non-humanoid heads in a single feed-forward pass. The resulting blendshapes support real-time animation and retargeting via a fixed expression vocabulary.
  }
\label{fig:ffr}
\end{figure*}

\input{sections/introduction}

\input{sections/related_works}

\input{sections/method}

\input{sections/experiments}

\section{Conclusion}
We presented RegHead, a novel framework for constructing semantic blendshape sets for non-humanoid head avatars under a fixed expression vocabulary. We build a large-scale non-humanoid head dataset, introduce a dense stochastic anchor motion representation, and propose a fast feed-forward registration model. Experiments show that our method improves expression fidelity over baselines while running orders of magnitude faster than optimization methods, and enables real-time retargeting from human face tracking to non-humanoid characters.

\paragraph{Limitations.}
RegHead can underfit identities far from common head anatomy, such as insect-like or long-tail species with ambiguous eyes, mouths, or facial regions. These cases have weaker semantic correspondence and may require different local motion patterns for the same expression label. The final blendshapes are also affected by artifacts or identity drift in the upstream unregistered meshes. 

\clearpage
\input{sections/appendix}


%
%
\bibliographystyle{splncs04}
\bibliography{main}
\end{document}

%% file: sections/introduction.tex
\section{Introduction}

The creation of animatable 3D non-humanoid head avatars is increasingly important for AR communication~\cite{kiuchi2025minimates, wang2025raise}, social media~\cite{chen2026snapmoji, pan2024expressive}, and games~\cite{kang2025user, kaiser2025get}, yet remains underexplored.
A practical animation interface for non-humanoid heads should support controllable expressions and efficient retargeting across diverse identities, species, and stylized appearances.
\emph{Semantic blendshapes} provide such an interface by defining a fixed \emph{expression vocabulary}, a small set of interpretable expression representations in full correspondence~\cite{ma2016semantically, lewis2014practice, li2010example}.
Given this vocabulary, animation reduces to predicting a low-dimensional weight vector over expressions and linearly blending the corresponding blendshapes, which naturally supports real-time control.
Moreover, because the same vocabulary is shared across identities, expression weights carry consistent semantic meaning, enabling cross-identity retargeting without solving a new correspondence problem at animation time.
Despite these advantages, constructing semantic blendshape sets for diverse non-humanoid heads requires substantial artist effort~\cite{li2010example, lewis2014practice} or heavy optimization-based processes~\cite{Chen2025v2m4, luo2025t2bs, sabathier2025lim}.

Building semantic non-humanoid blendshapes in a feed-forward manner is challenging due to several underexplored bottlenecks.
1) Current image/video generation and editing models~\cite{wu2025qwenimagetechnicalreport,brooks2023instructpix2pix,yang2024cogvideox,liu2024video} do not reliably produce the \emph{same} predefined expression (e.g., a specific ``frown'' or ``eye-closed'') across different identities, making it hard to obtain large-scale 2D observations with consistent semantic labels.
2) Modern 3D/4D generation methods~\cite{xie2024sv4d, yao2025sv4d, xiang2025trellis2, hunyuan3d2025hunyuan3d, ren2024l4gm, xu20254dgt} can produce high-quality expression-specific surfaces, but their outputs typically lack vertex correspondence across expressions, preventing direct use as a blendshape basis.
Recovering correspondence is possible with optimization-heavy non-rigid registration or per-instance shape matching, but these approaches are expensive and difficult to scale.
3) Non-humanoid facial expressions often involve highly localized, species-dependent deformations (e.g., jaw articulation, eyelids) that are poorly captured by generic global deformation models~\cite{zhang2026rigmo, liu2025riganything, ActionMesh2025, zhang2025one}.
As a result, the core challenge is not only generating expression-specific assets, but recovering \emph{semantically consistent correspondence} while preserving fine local deformations in a scalable pipeline.

RegHead addresses these challenges with three components: a large-scale
non-humanoid head dataset (Sec.~\ref{sec:dataset}), a dense stochastic-anchor
motion representation (Sec.~\ref{sec:motion_repr}), and a feed-forward
registration model (Sec.~\ref{sec:registration}). We begin by defining a
semantic expression vocabulary $\mathcal{E}$ using a small set of high-quality,
artist-rigged non-humanoid head assets, which provides consistent semantic
observations for blendshape construction. We then scale this vocabulary to many
identities by fine-tuning an image editing model~\cite{wu2025qwenimagetechnicalreport}.
The resulting expression-specific meshes
$\{\tilde{\mathcal{M}}^t\}_{t=0}^{T}$ are visually plausible but not in
vertex correspondence.

The key novelty of RegHead is to amortize correspondence recovery into a learned
feed-forward registration model, instead of solving a new per-instance optimization problem.
RegHead predicts the deformation from the neutral mesh to each target expression
in a single pass. To handle diverse non-humanoid facial topology and anatomy, we avoid predefined skeletons, bones, or templates, and introduce dense stochastic anchors as a topology-agnostic motion representation. To make the prediction reliable for both large-support motions and fine localized facial changes, RegHead combines a global matcher for coarse alignment with a structured local matcher for fine-grained correspondence refinement. The model is trained without ground-truth point correspondences or deformation fields, using rendered target consistency as supervision.

Experiments show that RegHead produces higher-fidelity expression meshes than
feed-forward and optimization-based baselines, while running orders of magnitude
faster than optimization methods. The resulting semantic blendshapes further
enable real-time retargeting from human face tracking to non-humanoid characters.

The main contributions of RegHead are:
\begin{itemize}
    \item We construct a large-scale dataset of $\sim$20k non-humanoid identities
    with a fixed semantic expression vocabulary, enabling scalable study of
    non-humanoid head blendshape generation.
    \item We introduce dense stochastic anchors, a topology-agnostic motion
    representation for localized non-humanoid facial deformation without
    predefined bones, skeletons, or shared templates.
    \item We propose a feed-forward registration model that converts
    unregistered expression meshes into corresponded semantic blendshapes in a
    single pass, trained without ground-truth correspondences or deformation
    fields.
    \item We demonstrate improved fidelity and speed over feed-forward and
    optimization-based baselines, and enable real-time retargeting from human
    face tracking to non-humanoid characters.
\end{itemize}

%% file: sections/related_works.tex
\section{Related Works}

\subsection{Head Avatars}
\label{sec:rw_head_avatar}

Recent advancements have extended human head avatar pipelines~\cite{gaussianavatars, flashavatar, chu2024generalizable, he2025lam, luo2025splatface, nguyen2026ffavatar} to stylized or toonified characters through text or style conditioning~\cite{sang2022agileavatar, perez2024styleavatar, song2024texttoon, yoon2024lego, wang2025headevolver}. Among these, TextToon~\cite{song2024texttoon} generates a drivable toonified head avatar from a monocular video and text style.
LeGO~\cite{yoon2024lego} generates a stylized 3D face model with a surface deformation network on a 3D morphable model (3DMM).
HeadEvolver~\cite{wang2025headevolver} generates stylized head avatars via template mesh deformation and 2D diffusion priors. However, these methods typically require parametric priors such as 3DMM. While effective for humanoids, these methods struggle when the target geometry deviates significantly from human anatomy. T2Bs~\cite{luo2025t2bs} animates character heads by aligning static assets to diffusion-generated videos via deformable 3DGS. While this bypasses some template constraints, it depends on a video-generation-and-alignment pipeline and optimization-heavy deformation baking. In contrast, we propose a fully feed-forward framework that avoids template-based anatomical constraints while achieving fast inference speed.

\subsection{4D Asset Reconstruction}
\label{sec:rw_4dgen}

 Traditional 4D asset reconstruction methods~\cite{jiang2024consistentd, zeng2024stag4d} primarily rely on Score Distillation Sampling (SDS) with priors from pretrained image and video diffusion models. Although subsequent works~\cite{ren2024l4gm, xie2024sv4d, liang2024diffusiond} have made significant progress in bypassing SDS, they mainly reconstruct 4D content using implicit representations such as NeRF~\cite{mildenhall2020nerf} or 3DGS~\cite{kerbl20233d}. These approaches generally don't explicitly enforce temporal correspondence across frames, which limits their applicability in industrial scenarios, including AR and gaming, where topology-consistent assets are often required. To address temporal correspondence issues, methods such as DreamMesh4D~\cite{li2024dreammesh4d} and V2M4~\cite{Chen2025v2m4} adopt mesh-based representations and recover topology-consistent geometry by coupling reconstruction with per-sequence alignment and refinement. While they achieve correspondence consistency, they rely on optimization-based pipelines, leading to substantial test-time computation. To improve efficiency, more recent works~\cite{shi2025drive, ActionMesh2025, chen2026motion, jiang2026mesh4d} propose feed-forward frameworks that maintain correspondence by predicting deformation fields over an anchor mesh. DriveAnyMesh~\cite{shi2025drive} extends 3DShape2VecSet~\cite{zhang20233dshape2vecset} to condition on temporally consistent multi-view frames and predicts deformations of anchor point clouds. ActionMesh~\cite{ActionMesh2025} introduces a two-stage network that first reconstructs structure-aligned meshes and then predicts deformations relative to an anchor mesh. These approaches achieve strong performance in both speed and quality. However, they primarily focus on articulated or whole-object motion and do not explicitly address the highly localized, fine-grained, and species-dependent deformations required for non-humanoid head animation. In contrast, our method is specifically designed to model such fine-grained deformations. We introduce a tailored motion representation along with global and local matching modules to effectively capture fine-grained non-humanoid facial movements while maintaining fast inference speed.

\subsection{Animation Representation}
\label{sec:rw_rigging}

In parallel with direct vertex-level deformation methods~\cite{Chen2025v2m4, shi2025drive, ActionMesh2025}, another line of research focuses on preparing animation-ready assets by predicting skeletons and skinning weights, enabling mesh animation through skeletal transformations. Methods such as RigAnything~\cite{liu2025riganything}, MagicArticulate~\cite{song2025magicarticulate}, and UniRig~\cite{zhang2025one} automatically generate skeleton structures and corresponding skinning weights for diverse 3D assets, making static meshes articulation-ready at scale. RigMo~\cite{zhang2026rigmo} further extends this direction by jointly predicting rig structures and motion from mesh sequences, facilitating generic character animation directly from data. However, the predicted skeletons in these methods are typically sparse and designed for coarse articulated motion. As a result, they are not well suited for non-humanoid head animation, which requires highly localized and fine-grained deformations. In contrast, our method adopts a motion representation based on a dense set of anchor points to drive mesh deformation, enabling more precise modeling of subtle and detailed movements.


%% file: sections/method.tex
\section{Methods}


Our goal is to generate a semantically defined blendshape set
$\{\mathcal{B}^t\}_{t=0}^{T}$ associated with a predefined expression vocabulary
$\mathcal{E}=\{e_t\}_{t=1}^{T}$.
All blendshapes share topology and vertex correspondence, enabling standard linear blendshape animation:
\begin{equation}
\mathbf{v}(\mathbf{w})=\mathbf{v}^0+\sum_{t=1}^{T} w_t(\mathbf{v}^t-\mathbf{v}^0),
\end{equation}
where $\mathbf{v}^t$ denotes the vertex positions of $\mathcal{B}^t$ and $\mathbf{w}$ are blend weights.

RegHead addresses this problem in three stages.
First, we construct semantically labeled expression observations for a predefined expression vocabulary and build a large-scale non-humanoid character expression dataset (Sec.~\ref{sec:dataset}).
Second, we define a stochastic anchor-based motion representation on the neutral shape, which provides an efficient and locally expressive parameterization for non-humanoid facial deformation (Sec.~\ref{sec:motion_repr}).
Third, given expression-specific but unregistered meshes $\{\tilde{\mathcal{M}}^t\}_{t=0}^{T}$, we propose a feed-forward registration network to predict anchor transformations and convert them into a corresponded blendshape basis $\{\mathcal{B}^t\}_{t=0}^{T}$ (Sec.~\ref{sec:registration}). We further show our training objective and retargeting application in Sec.~\ref{sec:loss} and Sec.~\ref{sec:retarget}.

\subsection{Expression vocabulary and dataset}
\label{sec:dataset}

A key challenge is obtaining expression observations with \emph{consistent semantics} across identities.
This challenge is particularly acute for non-humanoid heads: to our knowledge, there is no publicly available large-scale non-humanoid head blendshape dataset, and existing public 3D/4D assets~\cite{objaverseXL,li20214dcomplete, wu2023omniobject3d} provide limited coverage in both identities and expressions.
We therefore start by defining a fixed semantic expression vocabulary $\mathcal{E}$ using a small collection of high-quality, artist-rigged non-humanoid head assets.

Instead of relying on unconstrained video generation, we fine-tune an image editing model~\cite{wu2025qwenimagetechnicalreport} using paired renderings from the artist library, enabling controlled edits from a neutral reference into each target expression in $\mathcal{E}$.
Applied to text-to-image generations, this model produces semantically labeled expression image sets at scale, effectively expanding the artist-defined vocabulary from a small curated library to thousands of synthetic identities.

Given the neutral and edited expression images, we reconstruct expression-specific 3D surfaces using image-conditioned 3D/4D generation pipelines, with additional consistency heuristics to reduce inter-expression drift~\cite{xiang2025trellis2,hunyuan3d2025hunyuan3d}.
This yields a set of raw expression meshes $\{\tilde{\mathcal{M}}^t\}_{t=0}^{T}$ that preserve the intended semantics but are not yet in vertex correspondence.
These meshes serve as input to our feed-forward registration model (Sec.~\ref{sec:registration}), which converts them into the final corresponded blendshape basis $\{\mathcal{B}^t\}_{t=0}^{T}$.

We construct a dataset of approximately 20k non-humanoid identities, each with a labeled expression set in the same semantic vocabulary. 
To maintain quality at scale, we additionally employ human annotators to filter low-quality generations and reconstruction failures.
Please find more dataset details in the project website.



\begin{figure*}[t]
\centering
\includegraphics[width=\textwidth]{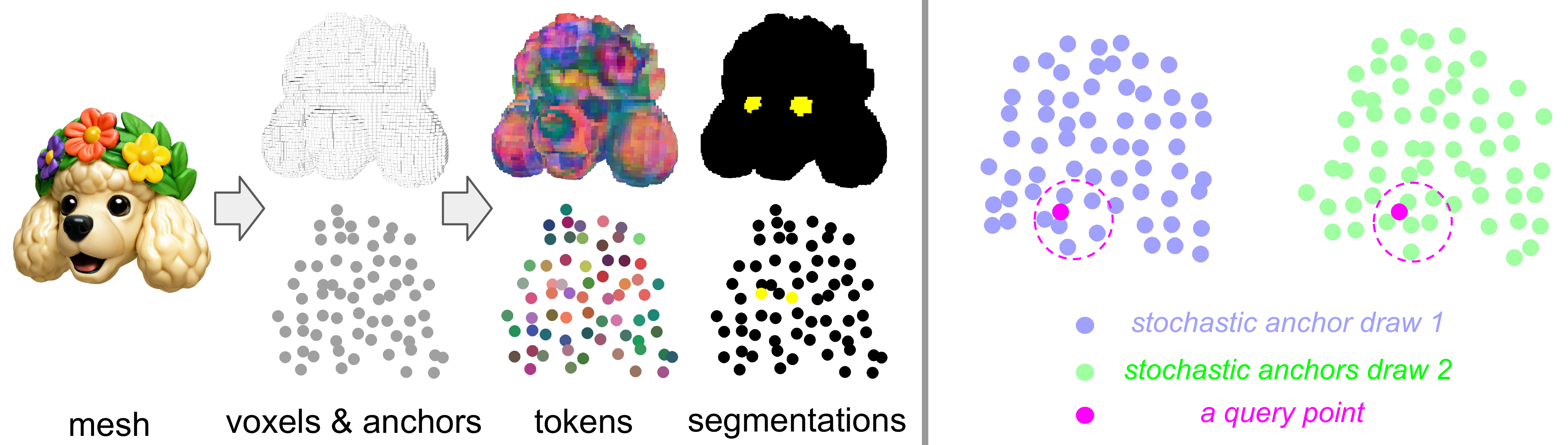}
\caption{
\textbf{(Left)} We voxelize the mesh and generate stochastic anchors, then create voxel tokens including segmentation cues unprojected from multi-view renderings of the mesh. We can also obtain per-anchor tokens or features by trilinear interpolation from voxels. \textbf{(Right)} Two draws of stochastic anchors (light purple and green) and a query point (magenta) driven by neighboring anchors. Although the identity of the neighboring anchors changes when generating stochastic anchors, the blended support remains localized.
  }
\label{fig:anchors}
\end{figure*}

\subsection{Stochastic Anchor Motion Representation}
\label{sec:motion_repr}

Our goal is to convert raw expression meshes $\{\tilde{\mathcal{M}}^t\}_{t=0}^{T}$ into a corresponded semantic blendshape basis $\{\mathcal{B}^t\}_{t=0}^{T}$ by predicting, for each expression $t$, an anchor-based deformation from the neutral mesh to the target.
Because non-humanoid facial motion is highly localized (e.g., eyelids) yet dense deformation on high-resolution surfaces is expensive, we represent motion as per-anchor transformations $\mathbf{T}^t$ defined on neutral anchors and propagate them to dense surface points via sparse-to-dense blending to obtain $\mathcal{B}^t$.
This subsection defines the motion representation (anchors, voxel tokens, and propagation), while Sec.~\ref{sec:registration} describes how $\mathbf{T}^t$ is predicted via correspondence-aware matching.

Let $\tilde{\mathcal{M}^0}$ denote the neutral mesh for an identity.
We sample a set of deformation anchors
$\mathbf{A}=\{\mathbf{a}_k\}_{k=1}^{K}$
on $\tilde{\mathcal{M}^0}$, where each anchor acts as a local control node.
A key design choice is that we \emph{resample anchors at every training iteration} and do not optimize anchor coordinates.
This stochastic anchor strategy improves local coverage and encourages the model to learn anchor-layout-invariant deformation prediction.
At test time we sample anchors from the same distribution.
Empirically, predictions are stable across different anchor draws. We find that using substantially denser anchor sets than seen during training does not improve fidelity and can introduce local artifacts due to a distribution shift in anchor spacing and blending weights.

We also sample a denser set of neutral query points
$\mathbf{Q}=\{\mathbf{q}_j\}_{j=1}^{N}$ from the neutral shape.
The model predicts transformations only at anchors, and propagates them to queries by blending nearby anchor motions, enabling efficient yet expressive dense deformation.

For each expression mesh $\tilde{{\mathcal{M}}^t}$, we voxelize it and build multi-view unprojection-based voxel tokens $\mathbf{V}^t$ inspired by prior voxel-latent pipelines~\cite{xiang2024structured}.
In our implementation, $\mathbf{V}^t$ concatenates an 8-channel voxel latent code with an unprojected semantic segmentation mask for the eye region, yielding 9-channel voxel tokens.
This formulation is modular: additional 2D cues can be unprojected into the same voxel grid and appended as extra channels when available.
Given any position, we obtain its feature under $\mathbf{V}^t$ by trilinear interpolation from neighboring voxels. We demonstrate this process in Fig.~\ref{fig:anchors} (left).

We realize the deformation function $f_\theta$ by predicting per-anchor \emph{pivoted similarity transforms}
$\mathbf{T}^t=\{\mathcal{T}_k^t\}_{k=1}^{K}$ for each target expression, which is obtained by:
\begin{equation}
\mathbf{T}^t = f_\theta(\mathbf{A}, \mathbf{V}^t).
\end{equation}

Then we propagate the transforms to queries via Linear Blend Skinning. For each query point $\mathbf{q}_j$, we precompute skinning weights $w_{jk}$ via normalized inverse Mahalanobis distances from its $K'$ nearest anchors, and deform it as
\begin{equation}
\hat{\mathbf{q}}_j^{\,t}
=
\sum_{k\in\mathcal{N}(j)} w_{jk}\,\phi(\mathbf{q}_j;\mathcal{T}_k^t,\mathbf{a}_k),
\qquad
\hat{\mathbf{Q}}^{\,t}=\{\hat{\mathbf{q}}_j^{\,t}\}_{j=1}^{M},
\end{equation}

where $\phi$ denotes applying the anchor transform to $\mathbf{q}_j$.

We demonstrate this process in Fig.~\ref{fig:anchors} (right). This sparse-to-dense propagation enables localized deformations while keeping the network prediction cost linear in the number of anchors.

\begin{figure*}[t]
\centering
\includegraphics[width=\textwidth]{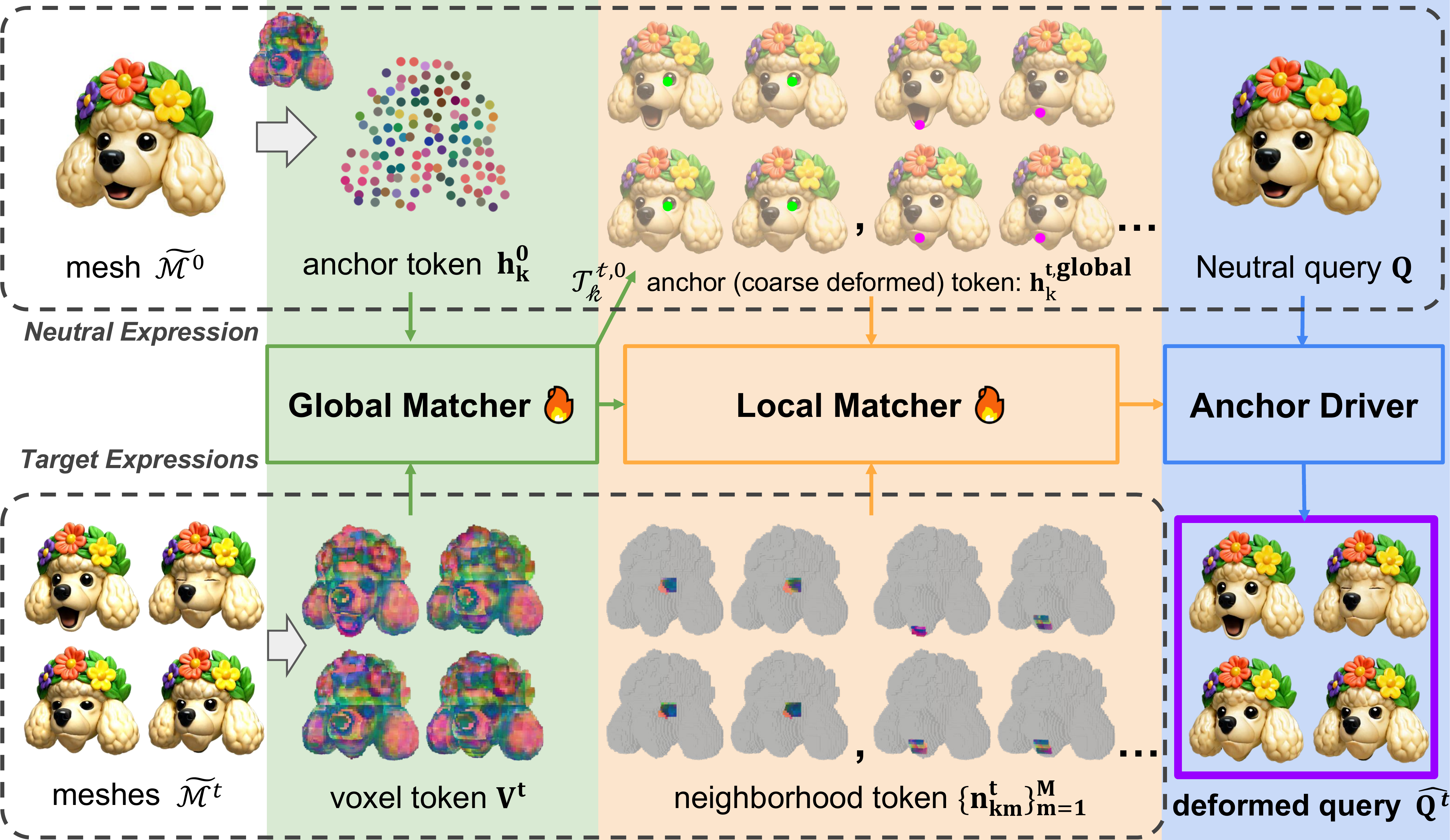}
\caption{
\textbf{Feed-Forward Registration.} The goal of our feed-forward registration is to learn a deformation function $f_\theta$ that deforms the neutral expression to multiple target expressions. We split all expressions of any identity into a neutral expression and target expressions and obtain anchor and voxel tokens as described in Sec.~\ref{sec:motion_repr}. Then, a global matcher first predicts a coarse deformation $\mathcal{T}_k^{t,0}$ to bring anchors into a better local matching basin, and a structured local matcher then refines fine-grained localized motion using validity-aware voxel neighborhoods. As marked in orange, we show two coarsely deformed anchors $\mathbf{a}_k^{t}$ (green and magenta) matching to their neighborhood tokens $\{\mathbf{n}_{km}^t\}_{m=1}^{M}$ shown at the bottom. Finally, the local matcher predicts the per-anchor transformations that drive the neutral query.
}
\label{fig:ffr}
\end{figure*}

\subsection{Feed-Forward Registration}
\label{sec:registration}

Given the anchors $\mathbf{A}$, query points $\mathbf{Q}$, and voxel tokens $\mathbf{V}^t$ defined in Sec.~\ref{sec:motion_repr}, we predict per-expression anchor transforms $\mathbf{T}^t$ using correspondence-aware coarse-to-fine matching.

Our core strategy is \emph{local feature matching} in voxel space: for each anchor, we compare its neutral token to voxel tokens within a nearby neighborhood of the target expression to estimate a local transformation.
This approach relies on the locality assumption that the true correspondence of an anchor lies within the queried neighborhood.
In practice, large-support motions (e.g., jaw articulation) can displace regions by more than the neighborhood radius, causing an anchor to retrieve neighbors from an unrelated facial region.
Moreover, voxel tokens are expression-dependent and may be affected by reconstruction artifacts, increasing ambiguity even when the neighborhood is correct.
To make local matching reliable, we adopt a coarse-to-fine design:
a global matcher first predicts a coarse deformation to bring anchors into the correct basin, after which a structured local matcher refines fine-grained localized motion using validity-aware voxel neighborhoods.
We demonstrate our feed-forward registration method in Fig.~\ref{fig:ffr}.

We represent anchors and voxel neighborhoods as feature tokens to enable attention-based matching.
For each anchor $\mathbf{a}_k$, we form a neutral anchor token $\mathbf{h}_k^0\in\mathbb{R}^d$ by encoding its position together with the interpolated neutral voxel tokens.
For each target expression $t$, we additionally construct a set of \emph{global tokens} $\mathbf{v}^t$ by randomly subsampling the valid voxels from the target feature field $\mathbf{V}^t$ and encoding each voxel’s position and feature.
These global tokens provide long-range context for coarse alignment.

We first perform a global matching stage that aggregates long-range evidence from the global tokens into each anchor token via cross-attention:
\begin{equation}
\mathbf{h}_k^{t,\mathrm{global}} = \mathrm{XAttn}\!\left(\mathbf{h}_k^{0}, \mathbf{v}^t\right).
\end{equation}
From $\mathbf{h}_k^{t,\mathrm{global}}$, we predict a coarse anchor transform $\mathcal{T}_k^{t,0}$ and warp the anchors to obtain coarse aligned anchor positions $\{\mathbf{a}_k^{t}\}$.
This stage provides a robust initialization for local refinement by improving the neighborhood quality for subsequent matching.

For local refinement, we gather a structured stencil neighborhood around each coarsely warped anchor $\mathbf{a}_k^{t}$ in $\mathbf{V}^t$ and encode each neighbor voxel into a token $\mathbf{n}_{km}^t$ using its relative offset and voxel tokens.
Instead of selecting neighbors by $k$NN in voxel space, we use a structured multi-radius stencil centered at each coarsely warped anchor $\mathbf{a}_k^{t}$ in the voxel tokens $\mathbf{V}^t$. We sample offsets on thin shell bands at multiple radii and cap the number of offsets per radius to keep a fixed token budget. We further use the voxel validity mask to discard invalid locations, and optionally snap invalid stencil slots to nearby valid voxels to avoid empty neighborhoods. Then, we encode each neighbor voxel into a token using its relative offset and voxel tokens, yielding neighborhood tokens $\{\mathbf{n}_{km}^t\}_{m=1}^{M}$.
Finally, we refine each anchor token by local cross-attention:
\begin{equation}
\mathbf{h}_k^{t,\mathrm{loc}} = \mathrm{LocalXAttn}\!\left(\mathbf{h}_k^{t,\mathrm{global}}, \{\mathbf{n}_{km}^t\}_{m=1}^{M}\right),
\end{equation}
followed by lightweight anchor-graph self-attention to encourage spatially coherent motion.
Finally, we predict a residual transform $\Delta\mathcal{T}_k^{t}$ and compose it with the coarse transform:
$\mathcal{T}_k^{t} = \Delta\mathcal{T}_k^{t} \circ \mathcal{T}_k^{t,0}$.
As shown in Sec.~\ref{sec:ablation}, this structured local matching stage is the primary contributor to fine expression deformation.

\subsection{Training Objectives}
\label{sec:loss}

Our training objective supervises the predicted deformation without requiring explicit point-wise correspondence between the neutral and target expressions.
Given neutral query points $\mathbf{Q}=\{\mathbf{q}_j\}_{j=1}^{N}$ sampled on the neutral mesh and their predicted deformed positions
$\hat{\mathbf{Q}}^{\,t}=\{\hat{\mathbf{q}}_j^{\,t}\}_{j=1}^{N}$ for expression $t$ (Sec.~\ref{sec:motion_repr}),
we optimize a differentiable rendering loss that compares $\hat{\mathbf{Q}}^{\,t}$ with the target expression appearance, together with an auxiliary geometric loss that stabilizes the global matching stage.

For each expression mesh $\tilde{\mathcal{M}}^t$, we sample a dense set of near-surface target points
$\mathbf{P}^t=\{\mathbf{p}_i^t\}_{i=1}^{N_P}$.
Each point carries its RGB color and additional segmentation obtained by interpolating the corresponding voxel channels from $\mathbf{V}^t$ (Sec.~\ref{sec:motion_repr}).
For the neutral query points $\mathbf{Q}$, we assign point attributes once from the neutral feature field $\mathbf{V}^0$ and transport them to $\hat{\mathbf{Q}}^{\,t}$.

We supervise deformation using a point-based differentiable renderer.
Specifically, we render both the predicted points $\hat{\mathbf{Q}}^{\,t}$ and the target points $\mathbf{P}^t$ as isotropic Gaussian splats under randomly sampled camera views $\pi$.
Compared to mesh-based rendering supervision, this point-based formulation provides stable gradients.
We compute both an RGB rendering loss and a segmentation rendering loss:
\begin{align}
\mathcal{L}_{\mathrm{rend}}
&=
\mathbb{E}_{\pi \sim \mathcal{D}_{\mathrm{cam}}}
\Big[
\mathcal{L}_{\mathrm{rgb}}(\pi) + \lambda_{\mathrm{seg}}\,\mathcal{L}_{\mathrm{seg}}(\pi)
\Big], \\
\mathcal{L}_{\mathrm{rgb}}(\pi)
&=
\left\|\mathcal{R}_{\mathrm{rgb}}(\hat{\mathbf{Q}}^{\,t};\pi) - \mathcal{R}_{\mathrm{rgb}}(\mathbf{P}^{t};\pi)\right\|_2^2
\\
&+
\lambda_{\mathrm{lpips}}\,
\mathrm{LPIPS}\!\left(\mathcal{R}_{\mathrm{rgb}}(\hat{\mathbf{Q}}^{\,t};\pi),\,\mathcal{R}_{\mathrm{rgb}}(\mathbf{P}^{t};\pi)\right). \\
\mathcal{L}_{\mathrm{seg}}(\pi)
&=
\left\|\mathcal{R}_{\mathrm{seg}}(\hat{\mathbf{Q}}^{\,t};\pi) - \mathcal{R}_{\mathrm{seg}}(\mathbf{P}^{t};\pi)\right\|_2^2,
\end{align}
where $\mathcal{R}_{\mathrm{rgb}}$ and $\mathcal{R}_{\mathrm{seg}}$ render the RGB and segmentation channels, respectively. We only use feature loss (LPIPS) on RGB rendering.
All splats use a fixed rotation and opacity; we set the isotropic scale per point based on local point density to obtain stable supervision.

To encourage the global matcher to predict a geometrically meaningful coarse deformation, we additionally supervise the coarsely deformed anchor positions $\mathbf{A}^{t}=\{\mathbf{a}_k^{t}\}_{k=1}^{K}$ (Sec.~\ref{sec:registration}).
Since $\mathbf{P}^t$ is dense, we uniformly subsample a smaller set $\mathbf{P}_{\mathrm{sub}}^t \subset \mathbf{P}^t$ with the same number as the anchors and minimize a Chamfer distance:
\begin{equation}
\mathcal{L}_{\mathrm{coarse}}
=
\mathrm{CD}\!\left(\mathbf{A}^{t}, \mathbf{P}_{\mathrm{sub}}^t\right).
\end{equation}
This provides a direct geometric signal to the global stage and improves neighborhood quality for subsequent local matching. The full training loss is
\begin{equation}
\mathcal{L} = \mathcal{L}_{\mathrm{rend}} + \lambda_{\mathrm{coarse}}\,\mathcal{L}_{\mathrm{coarse}}.
\end{equation}

\subsection{Real-Time Expression Retargeting}
\label{sec:retarget}

Given a human video, we run an off-the-shelf face-tracking system to detect dense facial landmarks and fit an internal human-head 3DMM, yielding per-frame expression coefficients
$\mathbf{w}^{\mathrm{hum}} \in \mathbb{R}^{D}$ and head pose. We learn a small MLP $g_\psi$ that maps tracking coefficients to the weights of our non-humanoid blendshape vocabulary without per-frame optimization:
\begin{equation}
\mathbf{w}
=
\!g_\psi(\mathbf{w}^{\mathrm{hum}}).
\label{eq:retarget_mlp}
\end{equation}

We optimize $g_\psi$ once offline using a combination of supervision in the tracker coefficient space and a 3D geometric consistency loss on a human face mesh driven by the tracker coefficients. In addition to the tracked expressions, we apply the tracked pose to normalized blendshapes to obtain real-time retargeting.


%% file: sections/experiments.tex
\section{Experiments}

\subsection{Implementation Details}
\label{sec:impl}


We randomly sample $K{=}3000$ anchors on the neutral mesh surface at every training iteration.
We precompute $N{=}30000$ canonical query points on the neutral shape. We shuffle the order of query points each iteration during training.
Each query point is driven by its $K'{=}10$ nearest anchors with fixed skinning weights (Sec.~\ref{sec:motion_repr}). For each identity, we normalize and align all expression meshes $\{\tilde{\mathcal{M}}^t\}$ to a shared canonical coordinate system and voxelize them into a $64^3$ grid. 
To construct the 8-channel voxel latent, we follow Xiang et al~\cite{xiang2024structured}.
To obtain a segmentation mask for each voxel, we render each mesh from 15 near-frontal viewpoints and unproject the multi-view features back into the voxel grid (Sec.~\ref{sec:motion_repr}).

We implement our method in PyTorch and train with mixed precision on 32$\times$A100 (80GB) GPUs.
Our training split contains 10k identities and the test split contains 200 identities; each identity provides 7 expressions.
For each identity, we designate a fixed neutral expression.
In each training iteration, we sample one identity (local batch size 1) by randomly selecting three target expressions.
The model then predicts deformations from the neutral to these three targets.
We optimize the feed-forward registration model using AdamW ($\beta_1{=}0.9$, $\beta_2{=}0.999$, weight decay $10^{-4}$) with a learning rate of $1{\times}10^{-4}$. We train for 2000 epochs and observe convergence after approximately 1500 epochs.






\subsection{Baseline Comparison}
\label{sec:baseline}



We compare RegHead with representative methods that can be adapted to the task
of converting an unregistered expression mesh set into an animatable,
corresponded blendshape basis. Because direct prior work on semantic blendshape
construction for diverse non-humanoid heads is limited, we select baselines that operate on general 3D geometry or 4D mesh sequences and can handle diverse topology and shape variation. 

For a fair comparison, all methods are provided with the same per-expression
target meshes $\{\tilde{\mathcal{M}}^t\}$ as input. This isolates the
registration problem from differences in upstream 3D
generation. We evaluate whether the predictions match
the target mesh $\tilde{\mathcal{M}}^t$ under the same rendering
conditions.

\noindent\textbf{ActionMesh}~\cite{ActionMesh2025} is a strong feed-forward
general mesh animation baseline. Its second stage predicts deformations of a
reference shape from an input mesh sequence. We use the authors' official
second-stage implementation.

\noindent\textbf{V2M4}~\cite{Chen2025v2m4} is an optimization-based method that registers a reference mesh to per-frame meshes from monocular video. We use the official implementation.

\noindent\textbf{T2Bs}~\cite{luo2025t2bs} aligns a static 3D asset to generative video outputs to obtain registered geometry~\cite{luo2025t2bs}. We report results produced by the authors on our test set, using multi-view renderings of our expression meshes as the video input.

\begin{table*}[t]
\centering
\scriptsize
\setlength{\tabcolsep}{3.5pt}
\resizebox{\textwidth}{!}{
\begin{tabular}{lccc cccc c}
\toprule
& \multicolumn{3}{c}{Rendered Appearance} 
& \multicolumn{4}{c}{Visible Geometry} 
& Runtime \\
\cmidrule(lr){2-4} \cmidrule(lr){5-8} \cmidrule(lr){9-9}
Method 
& PSNR $\uparrow$ & SSIM $\uparrow$ & LPIPS $\downarrow$
& CD $\downarrow$ & D-Err. $\downarrow$ & N-Cons. $\uparrow$ & Sil. IoU $\uparrow$
& Time $\downarrow$ \\
\midrule
ActionMesh & 22.2731 & 0.8989 & 0.0654 
           & 0.00513 & 0.01032 & 0.9103 & 0.9739 
           & \textbf{$\sim$5\,s} \\
V2M4       & 22.7849 & 0.9056 & 0.0660 
           & 0.00959 & 0.01827 & 0.9039 & 0.9567 
           & $\sim$100\,s \\
T2Bs       & 23.4349 & 0.9064 & 0.0568 
           & 0.00387 & 0.00711 & 0.9160 & 0.9851 
           & $\sim$1800\,s \\
RegHead    & \textbf{23.8421} & \textbf{0.9078} & \textbf{0.0551} 
           & \textbf{0.00358} & \textbf{0.00613} & \textbf{0.9192} & \textbf{0.9865} 
           & \textbf{$\sim$5\,s} \\
\bottomrule
\end{tabular}
}
\caption{
\textbf{Quantitative comparison with baselines.}
We evaluate both rendered appearance and visible geometry. Rendered appearance is measured by PSNR, SSIM, and LPIPS under the same camera views. Visible geometry is computed on rasterized first-hit surfaces from multiple views, avoiding hidden/internal mesh surfaces. CD denotes visible Chamfer distance, D-Err. denotes mean absolute depth error, N-Cons. denotes normal consistency, and Sil. IoU denotes silhouette intersection-over-union. 
}
\label{tab:baseline_quant_all}
\end{table*}

\begin{figure*}[t]
\centering
\includegraphics[width=\textwidth]{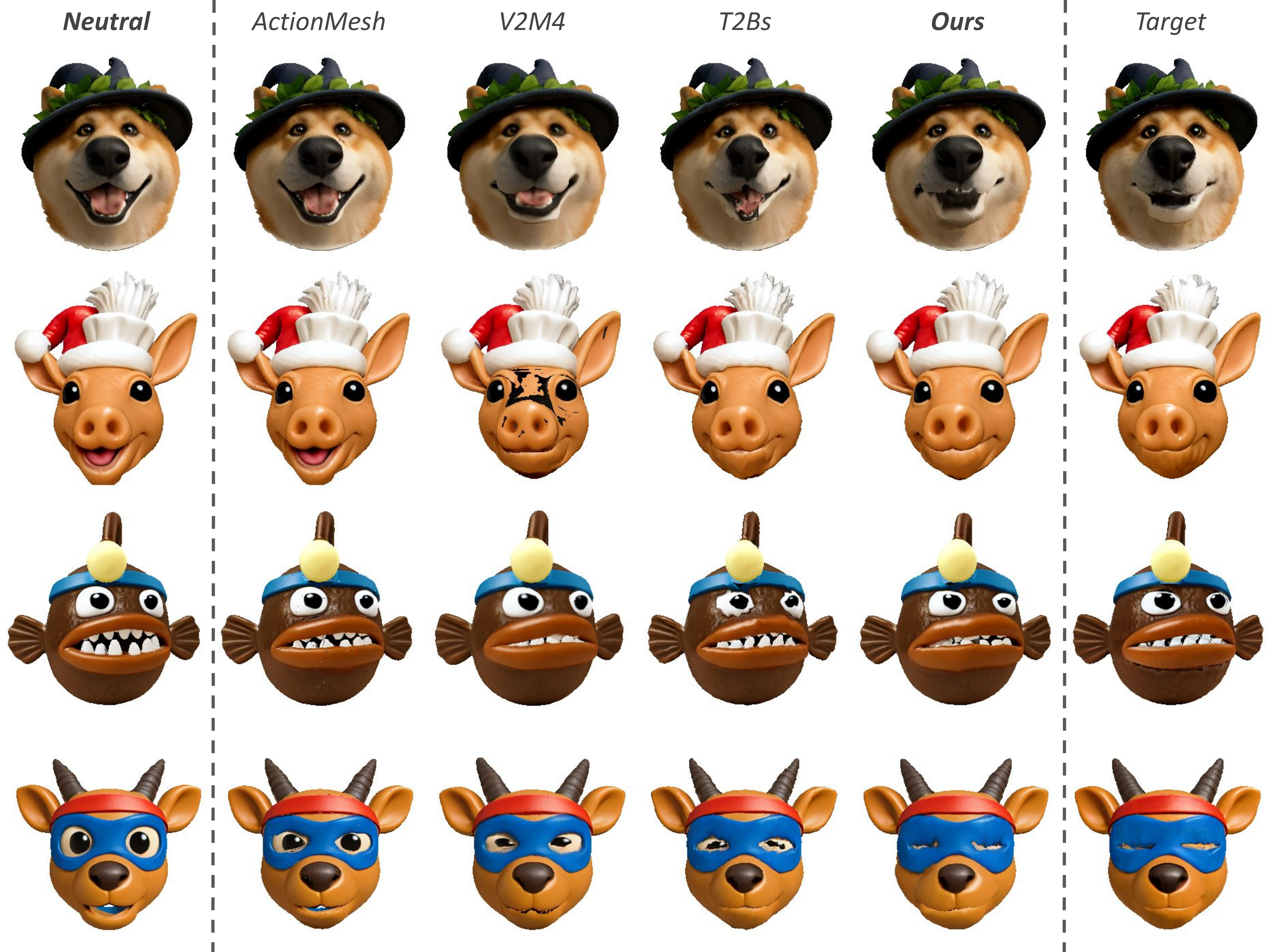}
\caption{
\textbf{Qualitative comparison with baselines.}
We show a subset of expressions from our 7-expression vocabulary.
Given the same neutral input, all methods deform the neutral shape to match four representative target expressions.
We render each method's predicted expression alongside the target unregistered mesh for visual reference.
Our method more faithfully reproduces localized expression changes while preserving overall identity and surface quality.
}
\label{fig:baseline}
\end{figure*}

\subsubsection{Quantitative Comparison}
\label{sec:baseline_quant}

We evaluate each method from two complementary perspectives. First, since the
output blendshapes are ultimately used for rendering and retargeting, we measure image-space expression fidelity by rendering each predicted expression mesh and the target mesh under the same camera views, and report PSNR/SSIM/LPIPS together with runtime per identity. 
Second, we evaluate visible-surface geometry. 
To handle coordinate differences across methods, we first estimate a global neutral-to-neutral alignment for each identity and method, and apply the same transform to all expressions. 
We rasterize each predicted expression and the target mesh from multiple views and compare only the first-hit visible surface. Specifically, we extract visible 3D points for Chamfer distance, compare z-buffer depth and surface normals on commonly visible pixels, and compute silhouette IoU from the rasterized masks. This avoids hidden or internal mesh surfaces and evaluates the geometry that contributes to the rendered avatar. Table~\ref{tab:baseline_quant_all} reports quantitative results on expression fidelity, visible geometry, and runtime. RegHead outperforms the feed-forward baseline ActionMesh by a clear margin. Compared to optimization-based methods, V2M4 and T2Bs, RegHead attains higher rendered fidelity and better visible-surface geometry while being orders of magnitude faster.

\subsubsection{Qualitative Comparison}

Fig.~\ref{fig:baseline} compares our predicted blendshapes with representative baselines on the same input neutral mesh and the same target expression meshes.
The selected expressions include both large-support motion (mouth closure) and subtle localized deformation (eyes), which are particularly challenging for non-humanoid heads. Overall, our method better captures fine-grained, spatially localized expression changes while maintaining coherent global structure.
ActionMesh often obviously underestimates localized facial motion, which is consistent with the absence of an explicit local matching mechanism in its deformation stage and with the domain gap between its training data and diverse non-humanoid head geometries.
Optimization-based methods, V2M4 and T2Bs, produce plausible global trends but frequently exhibit artifacts or underfitting on subtle regions such as partial eyelid closure, leading to visibly mismatched expressions under the same rendering conditions.

\begin{figure*}[t]
\centering
\includegraphics[width=\textwidth]{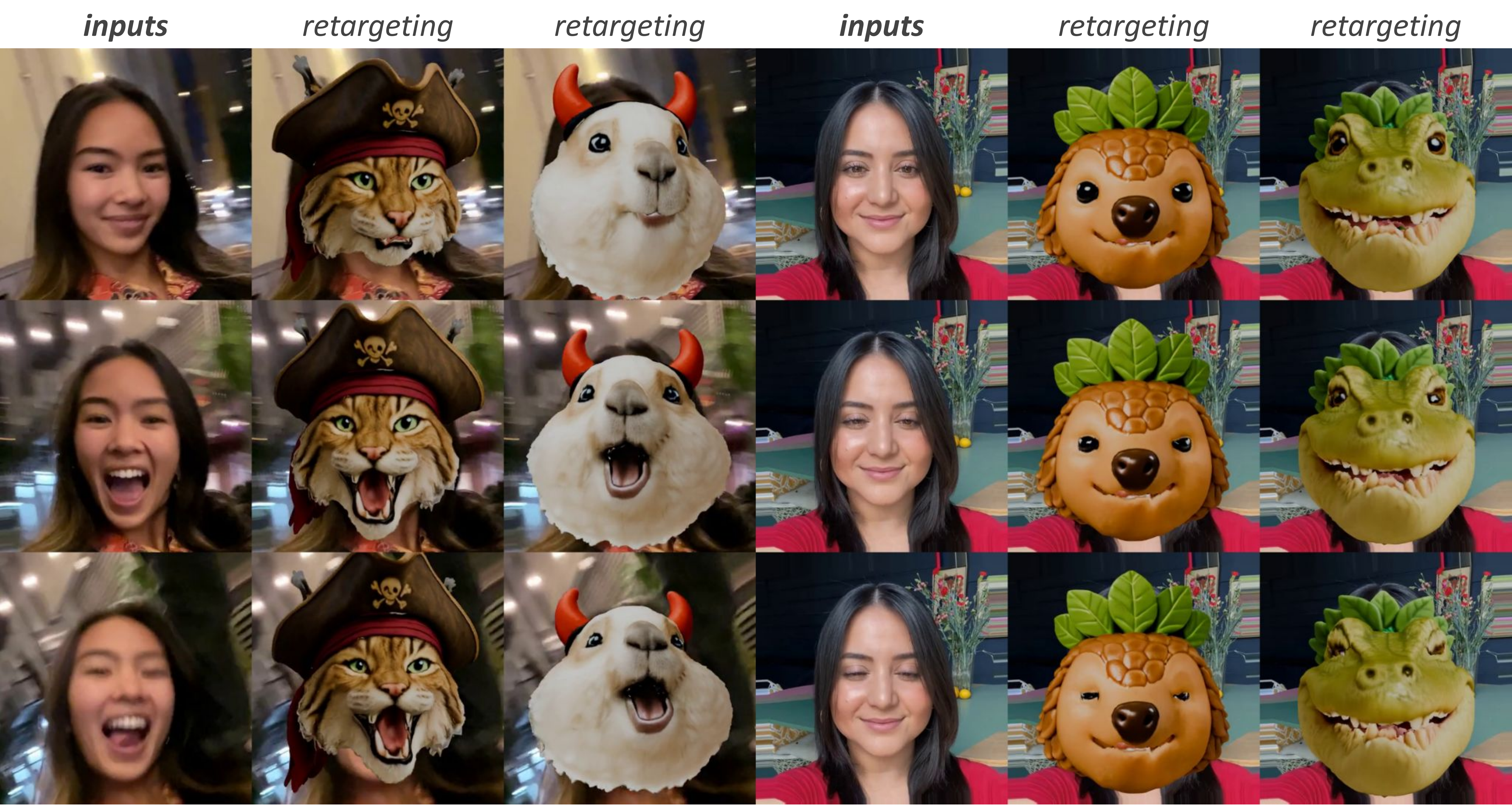}
\caption{
\textbf{Real-time retargeting from human performance to virtual characters.}
We show two human identities and four virtual characters driven by the same tracked performance.
We show pose and expression changes across each frame. 
The retargeted characters reproduce both head pose and localized facial motions.
}
\label{fig:retarget}
\end{figure*}


We show additional qualitative results that demonstrate real-time retargeting from human performance as described in Sec.~\ref{sec:retarget}, and animate non-humanoid characters via our blendshapes. Fig.~\ref{fig:retarget} shows retargeting results on two different human identities and four non-humanoid characters (two per identity).
We include sequences dominated by pose changes and localized expression changes. Across identities and characters, the retargeted virtual characters follow both the tracked head pose and the fine-grained facial expressions, demonstrating that our blendshape vocabulary provides a stable control interface for real-time animation.

\subsection{Ablation Study}
\label{sec:ablation}

We ablate key design choices in our motion representation (Sec.~\ref{sec:motion_repr}) and feed-forward registration model (Sec.~\ref{sec:registration}). See the supplement for more analysis.





\begin{table}[!t]
\centering
\small

\begin{subtable}[t]{0.49\textwidth}
\centering
\begin{tabular}{lccc}
\toprule
Method & PSNR$\uparrow$ & SSIM$\uparrow$ & LPIPS$\downarrow$ \\
\midrule
Ours & \textbf{23.842} & \textbf{0.9078} & \textbf{0.0551} \\
Fixed anchors & 23.323 & 0.9044 & 0.0610 \\
300 anchors & 22.226 & 0.8987 & 0.0629 \\
w/o seg. & 23.391 & 0.9056 & 0.0575 \\
\bottomrule
\end{tabular}
\caption{\textbf{Motion representation} (Sec.~\ref{sec:motion_repr}).}
\label{tab:ablation_repr}
\end{subtable}\hfill
\begin{subtable}[t]{0.49\textwidth}
\centering
\begin{tabular}{lccc}
\toprule
Method & PSNR$\uparrow$ & SSIM$\uparrow$ & LPIPS$\downarrow$ \\
\midrule
Ours & \textbf{23.842} & \textbf{0.9078} & \textbf{0.0551} \\
w/o local & 22.138 & 0.8963 & 0.0635 \\
w/o global & 23.660 & 0.9062 & 0.0577 \\
$k$NN neighbors & 23.055 & 0.9044 & 0.0574 \\
\bottomrule
\end{tabular}
\caption{\textbf{Registration design} (Sec.~\ref{sec:registration}).}
\label{tab:ablation_reg}
\end{subtable}
\caption{\textbf{Ablations} on stochastic anchor motion representation (a) and feed-forward registration design (b).}
\label{tab:ablations_two}
\end{table}

\paragraph{Ablations on stochastic anchor motion representation.}
Table~\ref{tab:ablation_repr} evaluates components of our motion representation.
Our full model uses $K{=}3000$ stochastic shell anchors and additionally unprojects semantic segmentation cues into the voxel feature field.
Replacing stochastic anchors with a fixed anchor set degrades performance with $-0.52$ PSNR and higher LPIPS, supporting the benefit of anchor-layout invariance induced by resampling.
Reducing the anchor count to $K{=}300$ leads to a large drop across all metrics, indicating that dense anchors are important for capturing fine-scale facial motion.
Finally, removing the unprojected segmentation channel (\emph{w/o seg.}) also consistently worsens results, demonstrating that fusing lightweight 2D semantic cues into the voxel field provides useful region-aware guidance.

\paragraph{Ablations on feed-forward registration.}
Table~\ref{tab:ablation_reg} evaluates the two-stage coarse-to-fine registration design.
Removing the local matching stage (\emph{w/o local}) causes a substantial degradation, confirming that local feature matching is the primary driver of fine expression deformation.
Removing the global alignment stage (\emph{w/o global}) yields a smaller but consistent drop, suggesting that the Global matcher improves robustness by placing anchors into a better basin for subsequent local matching.
Finally, replacing our validity-aware stencil neighborhood with a $k$NN neighborhood of the same token budget (the same number of neighbors) reduces performance, validating the benefit of structured multi-radius stencil neighborhoods for stable and informative local matching.


\begin{figure}[t]
\includegraphics[width=\linewidth]{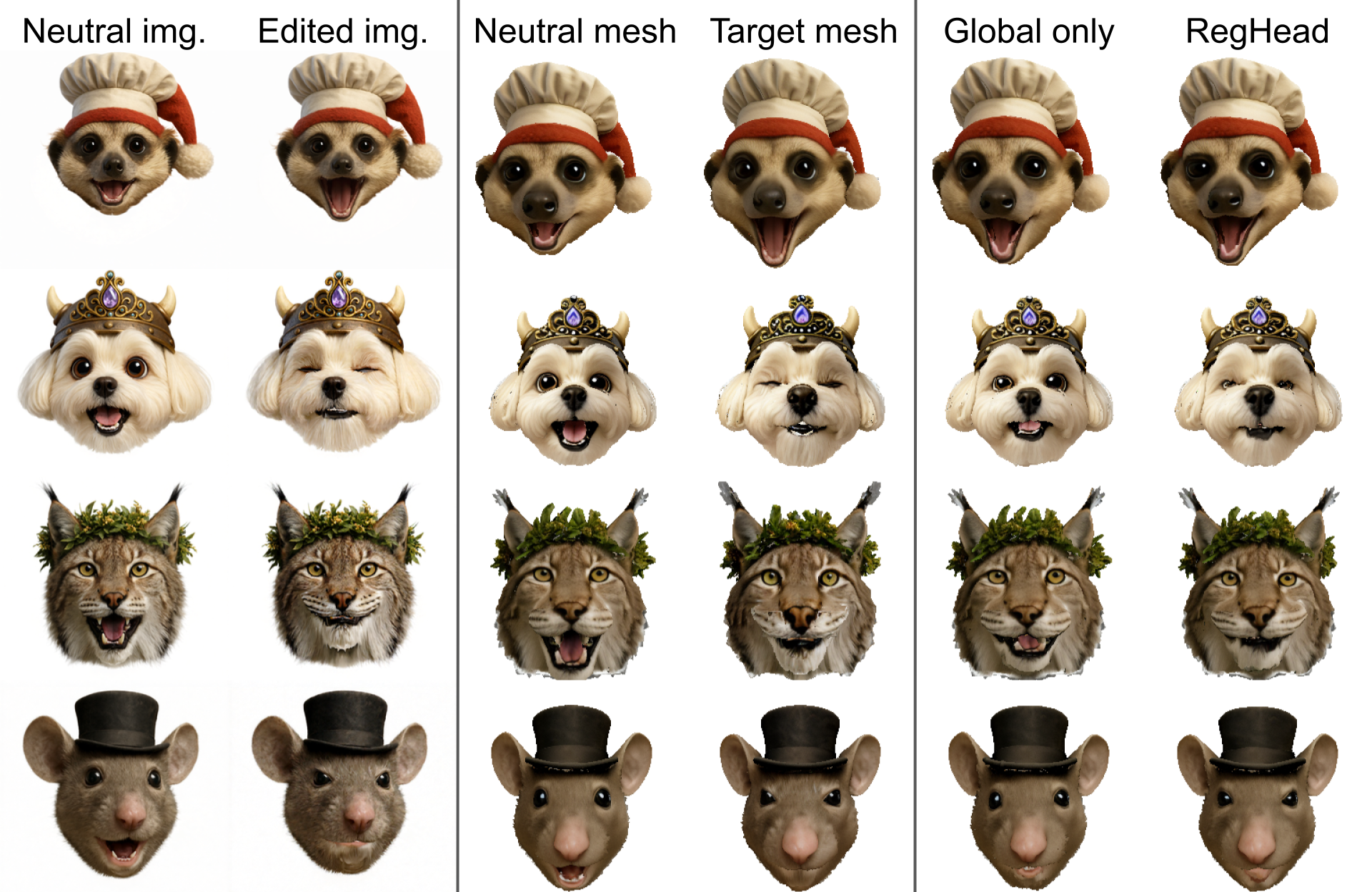}
\caption{
\textbf{Stage-wise diagnosis of expression generation.}
Columns show the neutral image, edited target image, raw neutral mesh,
raw target mesh, deformation using only the global matcher, and final
RegHead result. The global matcher captures coarse expression motion,
while the full model recovers stronger localized deformation.
}
\label{fig:stagewise_xsqm}
\end{figure}

\paragraph{Stage-wise Diagnosis of Expression Generation.} We provide a stage-wise visualization in Fig.~\ref{fig:stagewise_xsqm}. For each example, we show the neutral image, edited target image, raw neutral mesh, raw target expression mesh, the deformation predicted using only the
global matcher, and the final RegHead result.

This analysis separates two sources of expression variation. First, the edited
images and raw target meshes reveal the amplitude and locality of the upstream
expression targets. Some expressions are naturally concentrated in small
semantic regions, such as eyelids or mouth corners, rather than producing large
global deformation. Second, comparing the global-matcher-only output with the
final RegHead result shows the effect of our coarse-to-fine registration design.
The global matcher captures the coarse direction of motion, but often
underfits localized details. The full model, with structured local matching,
recovers stronger local deformation and better matches the raw target mesh.
This suggests that the remaining subtle cases are largely related to localized or moderate upstream targets and the difficulty of matching fine non-humanoid facial motion.

%% file: sections/appendix.tex
\begin{center}
    \large\bfseries Appendix / Supplementary materials
\end{center}

\section{Dataset Details}
\label{sec:s1}

More detailed dataset details are available on the project page and GitHub.

\begin{figure*}[!b]
\centering
\includegraphics[width=\textwidth]{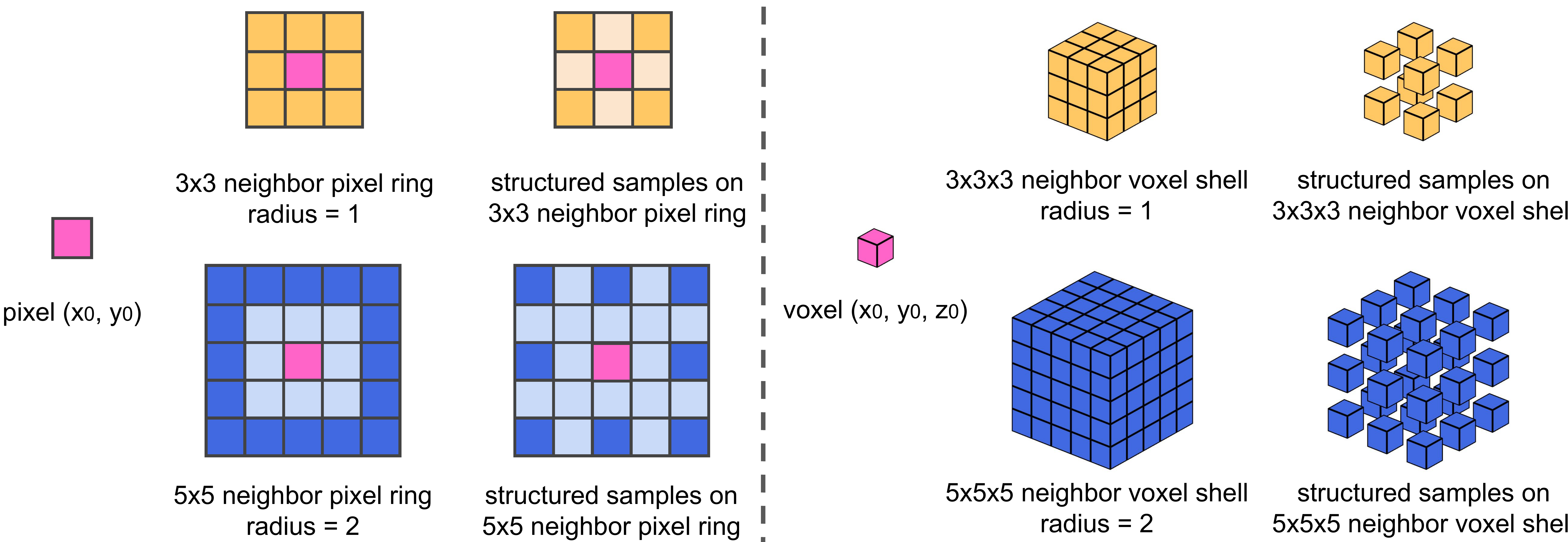}
\caption{\textbf{Visualization of structured stencil neighborhoods in 2D and 3D.} \\
\textbf{Left:} 2D analogy. For a center pixel, we consider ring (edge) neighborhoods at radii $r{=}1$ (3$\times$3) and $r{=}2$ (5$\times$5), and select a fixed set of structured samples on each ring to ensure uniform coverage with a fixed budget.
\textbf{Right:} 3D voxel stencil used in our local matcher. For a center voxel, we consider shell neighborhoods at multiple radii (e.g., 3$\times$3$\times$3 and 5$\times$5$\times$5 shells) and sample a capped set of offsets per radius.
}
\label{fig:s3}
\vspace{-15pt}
\end{figure*}

\section{Structured stencil neighborhood (feed-forward registration).}
\label{sec: stencil}

For each coarsely warped anchor position $\mathbf{a}_k^{t}$, we construct a fixed-budget set of neighborhood voxel tokens from the target voxel field $\mathbf{V}^t$.
We first convert $\mathbf{a}_k^{t}$ to voxel-grid coordinates and take the base index $(x_0,y_0,z_0)$.
Rather than selecting neighbors by $k$NN, we use a \emph{multi-radius stencil}: we predefine integer offsets on thin shell bands at radii $r\in\{1,2,3\}$ and cap the number of offsets per radius to keep a fixed token budget. Specifically, we use 8 offsets at $r{=}1$ and 26 offsets at $r{=}2$ and $r{=}3$.
Each offset produces a candidate voxel $(x,y,z)=(x_0,y_0,z_0)+\Delta$.
We illustrate the same idea in Fig.~\ref{fig:s3}: the left panel shows a 2D analogy where we sample structured points on ring (edge) neighborhoods, and the right panel shows our 3D voxel stencil that samples offsets on multi-radius shell bands.
This visualization highlights how the stencil provides multi-scale, well-distributed local context under a fixed token budget.

We then apply a voxel validity mask to discard empty locations and optionally \emph{snap} invalid stencil slots to the nearest valid voxel within a small local cube (radius $r_{\mathrm{snap}}$), ensuring non-empty neighborhoods even near boundaries.
Finally, for each selected voxel center $\mathbf{x}_{km}$ we form a neighborhood token by encoding its relative offset $\mathbf{x}_{km}-\mathbf{a}_k^{t}$ together with voxel features and feature differences to the neutral field, yielding tokens $\{\mathbf{n}_{km}^{t}\}$ used for local cross-attention.

\section{Additional Ablation Study}
\label{supsec:more_results}

\subsection{Local Amplitude Analysis}

We further quantify whether the predicted blendshapes preserve the local
amplitude of the target expressions. This is important because our expression
vocabulary is designed for natural semantic rig controls, rather than
maximally exaggerated expressions. More dramatic controls can be added by
expanding the training targets, for example by using stronger artist-designed
rigs to fine-tune the image-editing model and generate more expressive 3D
targets. Here, we focus on whether each method preserves the amplitude of the
current target expressions.

For each identity and expression, we render the neutral and target meshes from
multiple views and use SAM3 to identify semantic eye or mouth regions. The
active region is defined from the target pair, i.e., the local image-space
changes between the GT neutral and GT target meshes inside the SAM3 semantic
region. We then measure the target amplitude
$A^{\mathrm{target}}$ as the average depth/RGB change between the GT neutral
and GT target, and the predicted amplitude $A^{\mathrm{pred}}$ as the
corresponding change between the method's neutral and predicted expression.
The reported value is the percentage of target amplitude preserved,
$100 \times A^{\mathrm{pred}} / A^{\mathrm{target}}$, where values closer to
$100\%$ indicate better preservation. Our implementation computes these
statistics over rendered views using SAM3-defined semantic masks and supports
per-metric as well as composite amplitude measurements. 

\begin{table}[t]
\centering
\scriptsize
\setlength{\tabcolsep}{3pt}
\resizebox{\linewidth}{!}{
\begin{tabular}{llccccc}
\toprule
Region & Expression & \textbf{Target amp.} & RegHead & ActionMesh & T2Bs & V2M4 \\
\midrule
Eye
& closed eyes       & 0.133 & \textbf{90.5\%}  & 42.9\% & 53.4\% & 87.2\% \\
Eye
& half-open eyes    & 0.096 & \textbf{86.5\%}  & 42.9\% & 52.5\% & 118.2\% \\
\midrule
Mouth
& frown             & 0.057 & 85.1\%  & 51.5\% & \textbf{90.9\%} & 80.7\% \\
Mouth
& smile             & 0.069 & \textbf{87.3\%}  & 72.9\% & 54.8\% & 83.2\% \\
Mouth
& half-open mouth   & 0.119 & \textbf{93.6\%}  & 43.3\% & 83.7\% & 112.5\% \\
Mouth
& open mouth        & 0.132 & \textbf{103.1\%} & 40.2\% & 89.6\% & 117.7\% \\
\bottomrule
\end{tabular}
}
\vspace{1mm}
\caption{
\textbf{Local amplitude analysis.}
We measure image-space depth/RGB changes inside SAM3-defined semantic regions.
\textbf{Target amp.} denotes the local change from the reference expression
\emph{closed\_mouth\_open\_eyes}. Method columns report the percentage of target
amplitude preserved, $100 \times A^{\mathrm{pred}}/A^{\mathrm{target}}$; values
closer to $100\%$ indicate better preservation. Target amplitudes are
expression-dependent, while RegHead preserves both moderate and large local
amplitudes consistently.
}
\label{tab:local_amp}
\vspace{-7.5mm}
\end{table}

Tab.~\ref{tab:local_amp} shows that target amplitudes vary substantially across
expressions and regions. In the mouth region, open-mouth expressions have
roughly twice the target amplitude of frown or smile, while in the eye region,
closed eyes have larger amplitude than half-open eyes. RegHead preserves both
moderate and large target amplitudes well, with percentages close to $100\%$
across most expressions. This supports the stage-wise observation above:
perceived subtlety is often due to localized or moderate target expressions,
rather than collapse of the registration stage. We also observe that highly
localized smile and frown motions remain more challenging, suggesting a useful
direction for expanding the artist-designed target set.

\subsection{Robustness to Stochastic Anchor Sampling}
\label{supsec:anchor_robust}

Our motion representation resamples deformation anchors during training (Sec.3.2), raising a natural question: \emph{how sensitive is inference to the random anchor draw?}
To answer this, we evaluate the same trained model while varying only the random seed used to sample stochastic anchors at test time.
Table~\ref{tab:sup_anchor_seed} shows that performance is highly stable across seeds, with only negligible variation in PSNR and essentially identical SSIM/LPIPS.
This supports our claim that the model does not overfit to a particular anchor layout and learns anchor-layout-invariant deformation prediction.

\begin{table}[t]
\centering
\small
\begin{tabular}{lccc}
\toprule
Setting & PSNR$\uparrow$ & SSIM$\uparrow$ & LPIPS$\downarrow$ \\
\midrule
Seed 0 & 23.854 & 0.9078 & 0.0551 \\
Seed 1 & 23.824 & 0.9078 & 0.0551 \\
Seed 2 & 23.829 & 0.9078 & 0.0551 \\
Seed 3 & 23.845 & 0.9078 & 0.0551 \\
Seed 4 (main) & 23.842 & 0.9078 & 0.0551 \\
\bottomrule
\end{tabular}
\caption{\textbf{Inference robustness to anchor draw.} We vary only the random seed used to sample stochastic anchors at test time. Results are stable across seeds; we report Seed~4 in the main paper.}
\label{tab:sup_anchor_seed}
\vspace{-17.5pt}
\end{table}

\subsection{Effect of Increasing Anchor Density at Inference}
\label{supsec:anchor_density_infer}

A second question is whether inference can benefit from using many more anchors once gradients are disabled.
Intuitively, denser anchors could provide finer spatial support, but they also change the anchor spacing and create a distribution shift relative to training (which uses $K{=}3000$).
Table~\ref{tab:sup_anchor_more} shows that increasing the anchor count at test time does not improve fidelity and in fact degrades all metrics, with the degradation becoming more pronounced at very large $K$.
We observe that overly dense anchor sets can introduce local artifacts, consistent with the model being optimized for a fixed anchor budget and neighborhood statistics during training.
For this reason, we use the same anchor sampling strategy and anchor count at test time as during training.

\begin{table}[t]
\centering
\small
\begin{tabular}{lccc}
\toprule
Setting & PSNR$\uparrow$ & SSIM$\uparrow$ & LPIPS$\downarrow$ \\
\midrule
$K{=}3000$ (train/infer) & 23.842 & 0.9078 & 0.0551 \\
$K{=}10000$ (infer only) & 23.704 & 0.9056 & 0.0561 \\
$K{=}30000$ (infer only) & 23.427 & 0.9022 & 0.0584 \\
\bottomrule
\end{tabular}
\caption{\textbf{Increasing anchor density at inference.} Using substantially more anchors than seen during training does not improve results and can degrade fidelity, likely due to a distribution shift in anchor spacing and blending neighborhoods.}
\label{tab:sup_anchor_more}
\vspace{-17.5pt}
\end{table}

\begin{figure*}[h!]
\centering
\includegraphics[width=\textwidth]{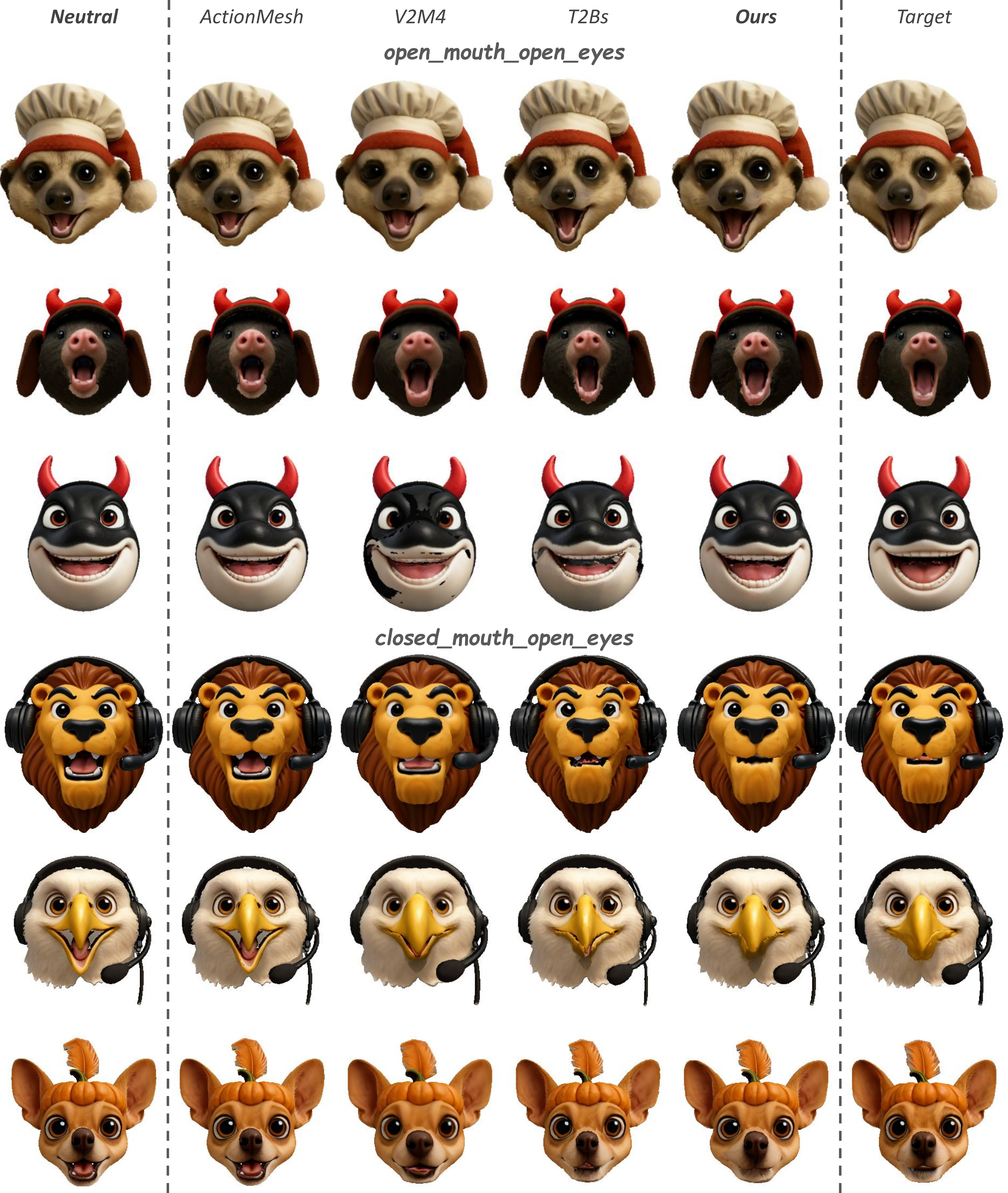}
\caption{
Qualitative comparison in addition to Fig. 4 of the main paper. Target expressions: \texttt{open\_mouth\_open\_eyes} and \texttt{closed\_mouth\_open\_eyes}.
}
\label{fig:s4_1}
\end{figure*}

\begin{figure*}[h!]
\centering
\includegraphics[width=\textwidth]{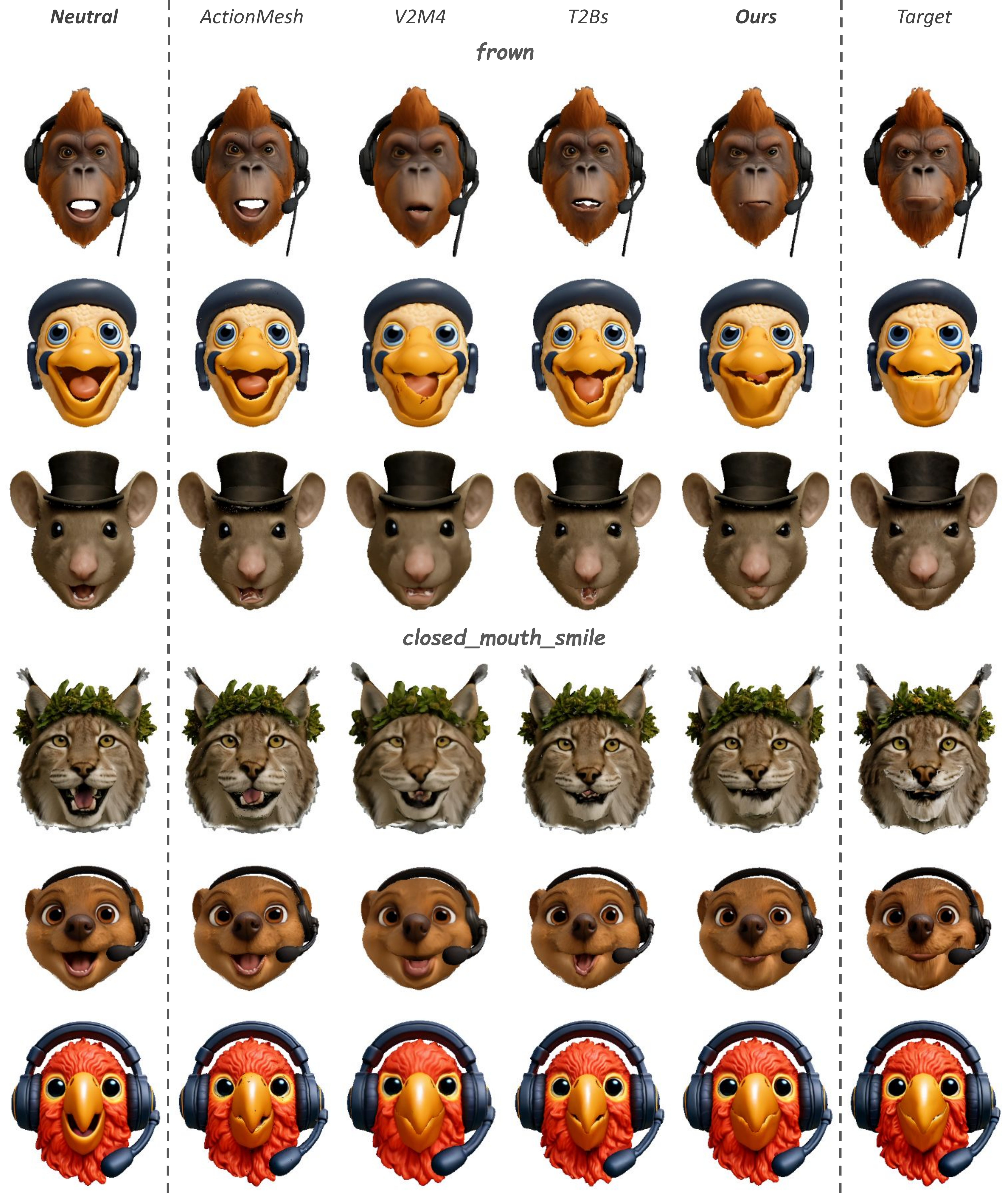}
\caption{
Qualitative comparison in addition to Fig. 4 of the main paper. Target expressions: \texttt{frown} and \texttt{closed\_mouth\_smile}.
}
\label{fig:s4_2}
\end{figure*}

\subsection{More Qualitative Comparison}
\label{supsec:more_qual}

Due to page limits, the main paper visualizes qualitative comparisons on a limited subset of identities and expressions.
In this supplementary section, we provide additional qualitative results for the 7-expression vocabulary.
For each expression, we show results on three representative identities and compare our predicted blendshapes against the same baselines used in Sec.~4.2, under identical rendering conditions.
Overall, these examples further confirm the trends observed in the main paper: our method more faithfully reproduces localized non-humanoid facial deformations (e.g., eyelids and subtle mouth motion) while maintaining coherent global structure.



\begin{figure*}[p]
\centering
\includegraphics[width=\textwidth]{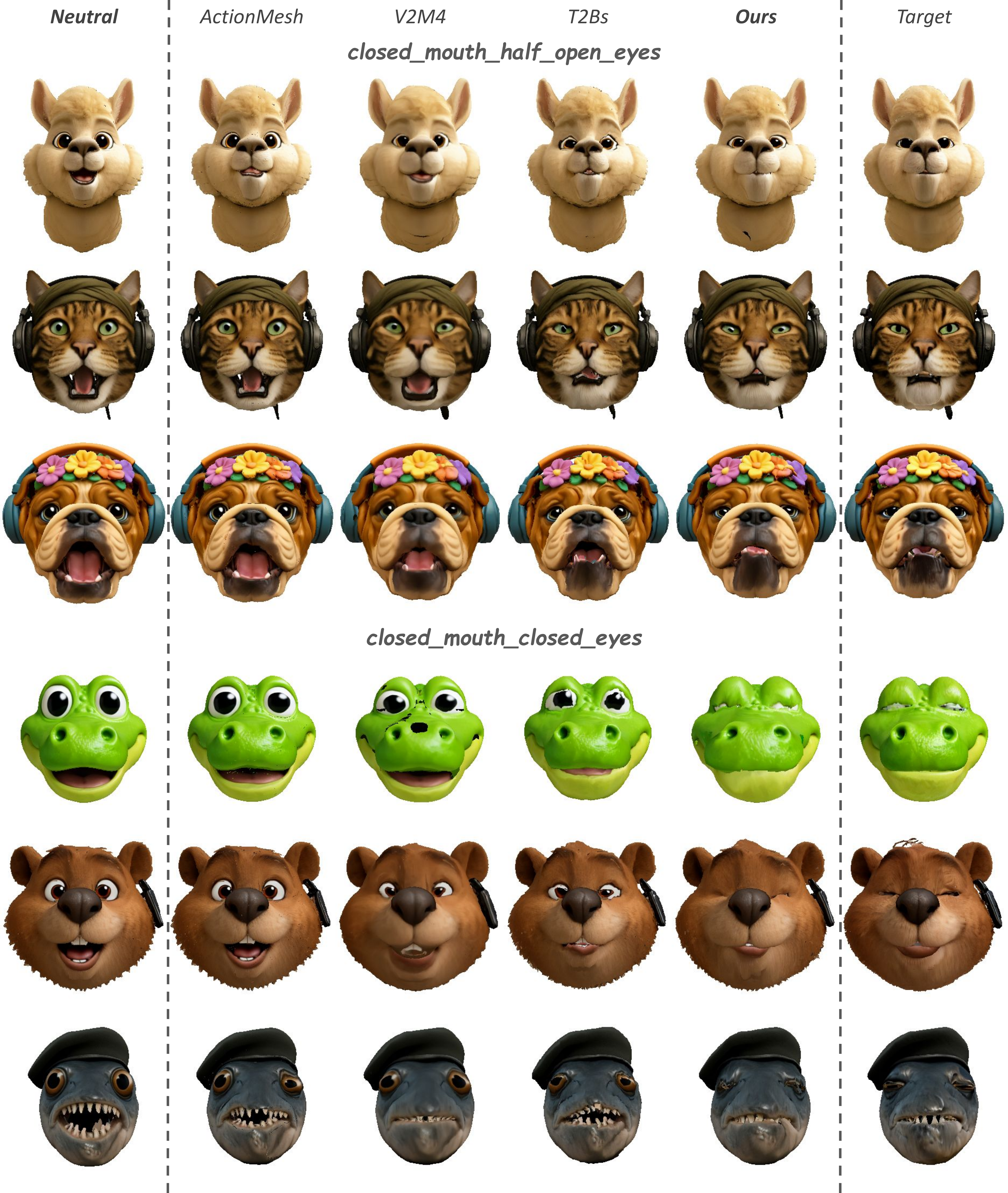}
\caption{
Qualitative comparison in addition to Fig. 4 of the main paper. Target expressions: \texttt{closed\_mouth\_half\_open\_eyes} and \texttt{closed\_mouth\_closed\_eyes}.
}
\label{fig:s4_3}
\end{figure*}



\clearpage